\documentclass[journal]{IEEEtran}
\usepackage{epsfig}
\usepackage{array}
\usepackage{hyperref}
\usepackage{multirow}
\usepackage{booktabs}

\makeatletter

\newcommand{\Rmnum}[1]{\expandafter\@slowromancap\romannumeral #1@}
\makeatother
\ifCLASSINFOpdf
\else
\fi
\usepackage{amsmath}
\usepackage{url}

\begin{document}
%
\title{RA-UNet: A hybrid deep attention-aware network to extract liver and tumor in CT scans}
%
%
%

\author{Qiangguo~Jin,
\and
Zhaopeng~Meng,
\and
Changming~Sun,
\and
Leyi~Wei,
\and
and~Ran~Su

\thanks{Manuscript submitted Oct. 4, 2018. (Corresponding author: Leyi Wei and Ran Su).}
\thanks{Qiangguo Jin is with School of Computer Software, College of Intelligence and Computing, Tianjin University, Tianjin, China (e-mail: qgking@tju.edu.cn).}
\thanks{Zhaopeng Meng is with School of Computer Software, College of Intelligence and Computing, Tianjin University, Tianjin, China (e-mail: mengzp@tju.edu.cn).}
\thanks{Changming Sun is with CSIRO Data61, Sydney, Australia (e-mail: changming.sun@csiro.au).}
\thanks{Leyi Wei is with School of Computer Science and Technology, College of Intelligence and Computing, Tianjin University, Tianjin, China (e-mail: weileyi@tju.edu.cn)}
\thanks{Ran Su is with School of Computer Software, College of Intelligence and Computing, Tianjin University, Tianjin, China (e-mail: ran.su@tju.edu.cn).}
}

\markboth{Journal of \LaTeX\ Class Files October~2018}%
{Shell \MakeLowercase{\textit{et al.}}: Bare Demo of IEEEtran.cls for IEEE Journals}
%



\maketitle

\begin{abstract}
Automatic extraction of liver and tumor from CT volumes is a challenging task due to their heterogeneous and diffusive shapes. Recently, 2D and 3D deep convolutional neural networks have become popular in medical image segmentation tasks because of the utilization of large labeled datasets to learn hierarchical features. However, 3D networks have some drawbacks due to their high cost on computational resources. In this paper, we propose a 3D hybrid residual attention-aware segmentation method, named RA-UNet, to precisely extract the liver volume of interests (VOI) and segment tumors from the liver VOI. The proposed network has a basic architecture as a 3D U-Net which extracts contextual information combining low-level feature maps with high-level ones. Attention modules are stacked so that the attention-aware features change adaptively as the network goes ``very deep" and this is made possible by residual learning. This is the first work that an attention residual mechanism is used to process medical volumetric images. We evaluated our framework on the public MICCAI 2017 Liver Tumor Segmentation dataset and the 3DIRCADb dataset. The results show that our architecture outperforms other state-of-the-art methods. We also extend our RA-UNet to brain tumor segmentation on the BraTS2018 and BraTS2017 datasets, and the results indicate that RA-UNet achieves good performance on a brain tumor segmentation task as well.
\end{abstract}


%
\IEEEpeerreviewmaketitle

\section{Introduction}
\label{sec:intro}
\begin{figure}
\centering
\includegraphics[scale=0.45]{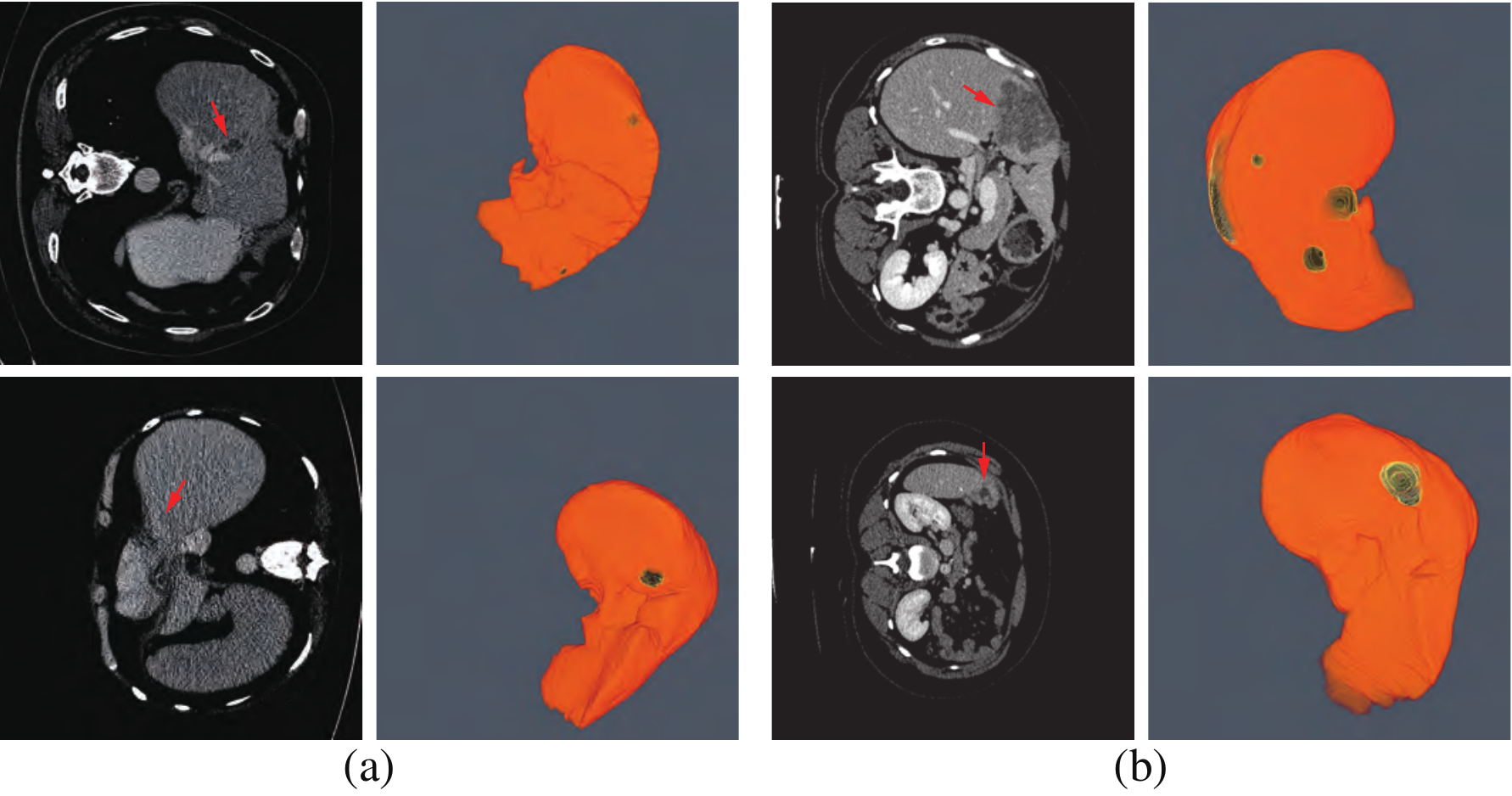}
\caption{Examples of typical 2D CT scans and 3D views of the corresponding ground truth of liver/tumor extractions where red arrows indicate the tumor/lesion regions. The orange regions denote the liver, and the darker regions within the yellow regions denote the tumors. (a) shows two slices from the MICCAI 2017 Liver Tumor Segmentation (LiTS) dataset. (b) shows two slices from the 3DIRCADb dataset.}
\label{fig:overview of dataset}
\end{figure}

%
%
%
%
\IEEEPARstart{L}{iver}  tumors, or hepatic tumors, are great threats to human health. The malignant tumor, also known as the liver cancer, is one of the most frequent internal malignancies worldwide (6\%), and is also one of the leading death causes from cancer (9\%)~\cite{whow_2014_1},~\cite{whow_2014_5}. Even the benign (non-cancerous) tumors sometimes grow large enough to cause health problems. Computed tomography (CT) is used to aid the diagnosis of liver tumors~\cite{christ2017automatic}. The extraction of liver and tumors from CT is a critical prior task before any surgical intervention in choosing an optimal approach for treatment. Accurate segmentation of liver and tumor from medical images provides their precise locations in the human body. Then therapies evaluated by the specialists can be provided to treat individual patients~\cite{rajagopal2015survey}. However, due to the heterogeneous and diffusive shapes of liver and tumor, segmenting them from the CT images is quite challenging. Numerous efforts have been taken to tackle the segmentation task on liver/tumors. Fig.~\ref{fig:overview of dataset} shows some typical liver and tumor CT scans.

 In general, the liver and tumor extraction approaches can be classified into three categories: manual segmentation, semi-automated segmentation, and automated segmentation. Manual segmentation is a subjective, poorly reproducible, and time-consuming approach. It heavily depends upon human recognizable features, and it requires people with high-level technical skills to carry out such tasks. These factors make it impractical for real applications~\cite{li2015automatic}. Semi-automated segmentation requires initial human intervention, which may cause bias and mistakes. In order to accelerate and facilitate diagnosis, therapy planning, monitoring, and finally help surgeons remove tumors, it is necessary to develop an automated and precise method to segment tumors from CT images. However, the large scale of spatial and structural variability, low contrast between liver and tumor regions, existence of noise, partial volume effects, complexity of 3D-spatial tumor features, or even the similarity of nearby organs make the automation of segmentation quite a difficult task~\cite{li2015automatic}. Recently, convolutional neural networks (CNN) have been applied to many volumetric image segmentations. A number of CNN models including both 2D and 3D networks have been developed. However, the 3D networks are usually not as efficient and flexible as the corresponding 2D networks. For instance, 2D and 3D fully convolutional networks (FCNs) have been proposed for semantic segmentation~\cite{long2015fully}. Yet due to the high computational cost and GPU memory consumption, the depth of the 3D FCNs is limited compared to that of 2D FCNs, which makes it impractical for 2D networks to be extended to 3D networks.

 To address these issues and inspired by the attention mechanism~\cite{vaswani2017attention} and the residual networks~\cite{he2016deep}, we propose a hybrid residual attention-aware liver and tumor extraction neural network named RA-UNet~\footnote{https://github.com/RanSuLab/RAUNet-tumor-segmentation.git}, which is designed to effectively extract 3D volumetric contextual features of liver and tumor from CT images in an end-to-end manner. The proposed network integrates a U-Net architecture and an attention residual learning mechanism which enables the optimization and performance improvement of very deep networks. To the best of our knowledge, this is the first work that attention residual mechanism is used in medical image segmentation tasks. The contributions of our works are listed as follows: Firstly, the residual blocks are stacked into our architecture which allows for a deeper architecture and can handle the gradient vanishing problem. Secondly, the attention mechanism can have the capability of focusing on specific parts of the image. Different types of attentions are possible through stacking attention modules so that the attention-aware features can change adaptively. Thirdly, we use the 2D/3D U-Net as the basic architecture to capture multi-scale attention information and to integrate low-level ones with high-level features. It is also worth noticing that our liver/tumor segmentation approach is a full 3D network which is used for the segmentation in an end-to-end fashion. Besides, our model does not depend on any pre-trained model or commonly used post processing techniques, such as 3D conditional random fields. The generalization of the proposed approach is demonstrated through testing on different datasets. Not only does our architecture extracts accurate liver and tumor regions but also achieves competitive performances comparing with other state-of-the-art methods on both the MICCAI 2017 Liver Tumor Segmentation (LiTS) dataset and the 3DIRCADb dataset~\cite{soler20103d}. Furthermore, we extend our RA-UNet to brain tumor segmentation tasks and it turned out that our RA-UNet is extendable to other medical image segmentation tasks. Our paper is organized as follows. In Section~\ref{sec:related_works}, we briefly review the current state-of-the-art automated liver tumor segmentation methods. We illustrate the methodology in details including the datasets, preprocessing strategy, hybrid deep learning architecture, and training procedure in Section~\ref{sec:methodology}. In Section~\ref{sec:exp_and_res}, we evaluate the proposed algorithm, report the experimental results, compare with some other approaches, and extend our approach to other medical segmentation tasks. Conclusions and future works are given in Section~\ref{sec:conclusion}.

\section{Related Works}
\label{sec:related_works}

Recently, deep neural networks (DNNs) have been used in a number of areas such as natural language processing and image analysis~\cite{liu2017survey}. Some have achieved state-of-the-art performance in medical imaging challenges~\cite{litjens2017survey}. Unlike the traditional methods that use hand-crafted features, DNNs are able to automatically learn discriminative features. The learned features which contain hierarchical information have the ability to represent each level of the input data. Among those methods, CNN is one of the most popular methods and has shown impressive performance for 3D medical image analysis tasks. Multi-scale patch-based and pixel-based strategies were proposed to improve the segmentation performance. For instance, Zhang et~al. proposed a method which used deep CNN for segmenting brain tissues using multi-modality magnetic resonance images (MRI)~\cite{zhang2015deep}. Li et~al. presented an automatic method based on 2D CNN to segment lesions from CT slices and compared the CNN model with other traditional machine learning techniques~\cite{li2015automatic}, which included AdaBoost~\cite{collins2002logistic}, random forests (RF)~\cite{breiman2001random} and support vector machine (SVM)~\cite{furey2000support}. This study showed that CNN still had limitations on segmenting tumors with uneven density and unclear borders. Pereira et~al. proposed a CNN architecture with small kernels for segmenting brain tumors on MRI data~\cite{pereira2016brain}. This architecture reached Dice similarity coefficient metrics of 0.78, 0.65, and 0.75 for the complete, core, and enhancing regions respectively. Lee et~al. presented a CNN-based architecture that could learn from provided labels to construct brain segmentation features~\cite{lee2011towards}. However, due to low memory requirements, low complexity of computation, and lots of pre-trained models, most of the latest CNN architectures including the methods reviewed above used 2D slices from 3D volumes for carrying out the segmentation task. However, the spatial structural organizations of organs are not considered and the volumetric information is not fully utilized. Therefore, 3D automatic segmentation which makes full use of spatial information is urgently needed for surgeons.

In order to sufficiently add 3D spatial structures into CNN for 3D medical image analysis, 3D CNN which considers axial direction of the 3D volumes has recently been put forward in the medical imaging field. Shakeri et~al. proposed a 2D CNN architecture to detect tumors from a set of brain slices~\cite{shakeri2016sub}. Then they additionally applied a 3D conditional random field (CRF) algorithm for post processing in order to impose volumetric homogeneity. This is one of the earliest studies that used CNN-related segmentation on volumetric images. {\c{C}}i{\c{c}}ek et~al. learned from sparsely sequential volumetric images by feeding U-Net with 2D sequential slices~\cite{cciccek20163d}. 3D CNN-based segmentation methods were then employed in a large scale. Andermatt et~al. used a 3D recurrent neural network (RNN) with gated recurrent units to segment gray and white matters in a brain MRI dataset~\cite{andermatt2016multi}. Dolz et~al. investigated a 3D FCN for subcortical brain structure segmentation in MRI images~\cite{dolz20173d}. They reduced the computational and memory costs, which was quite a severe issue for 3D CNN, via small kernels with a deeper network. Bui et~al. proposed a deep densely convolutional network for volumetric brain segmentation~\cite{bui20173d}. This architecture provided a dense connection between layers. They concatenated feature maps from fine and coarse blocks, which allowed to capture multi-scale contextual information. The 3D deeply supervised network (DSN), which had a much faster convergence and better discrimination capability, could be extended to other medical applications~\cite{dou20163d}. Oktay et~al. proposed a novel attention gate model called attention U-Net for medical imaging which could learn to concentrate on target structures of different shapes and sizes~\cite{oktay2018attention}. However, due to the hardware limitation, 3D convolutional medical image segmentation is still a bottleneck.

\begin{figure*}[ht]
\centering
\includegraphics[scale=0.55]{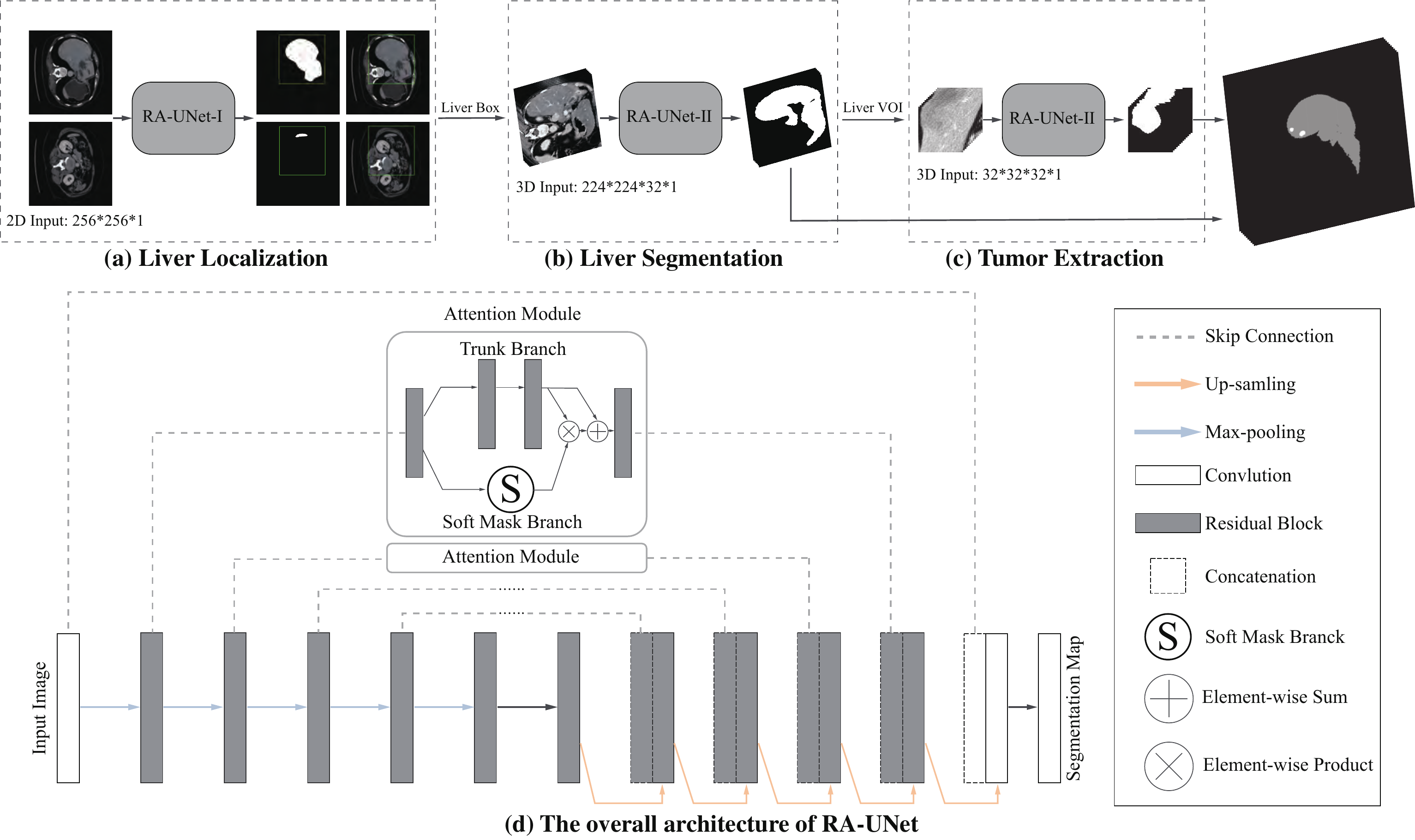}
\caption{Overview of the proposed pipeline of liver and tumor segmentation. (a) A simple version of 2D RA-UNet (RA-UNet-\Rmnum{1}) is employed for coarse localization of a liver region within a boundary box. (b) The 3D RA-UNet (RA-UNet-\Rmnum{2}) is designed for hierarchically extracting attention-aware features of liver VOIs inside the liver boundary box. (c) RA-UNet-\Rmnum{2} is responsible for an accurate tumor extraction which is inside the liver VOIs. (d) The overall architecture of RA-UNet.}
\label{fig:overview of architecture}
\end{figure*}

As for liver tumor detection in 3D volumetric images, not many explorations have been made using the CNN-based methods. Lu et~al. proposed a method based on 3D CNN to carry out the probabilistic segmentation task and used graph cut to refine the previous segmentation result. However, as tested only on one dataset, the generality of this architecture still needs to be validated~\cite{lu2017automatic}. Christ et~al. proposed a cascaded FCNs (CFCNs) to segment liver and its lesions in CT and MRI images, which enabled segmentation for large scale medical trials~\cite{christ2017automatic}. They trained the first FCN to segment the liver and trained the second FCN to segment its lesions based on the predicted liver region of interest (ROI). This approach reached a Dice score of 94\%. Additionally, Christ et~al. also predicted hepatocellular carcinoma (HCC) malignancy using two CNN architectures~\cite{christ2017survivalnet}. They took a CFCN as the first step to segment tumor lesions. Then they applied a 3D neural network called SurvivalNet to predict the lesions' malignancy. This method achieved an accuracy of 65\% with a Dice score of 69\% for lesion segmentation and an accuracy of 68\% for tumor malignancy detection. Kaluva et~al. proposed a fully automatic 2-stage cascaded method for liver and tumor segmentation based on the LiTS dataset, and they reached global Dice scores of 0.923 and 0.623 on liver and tumor respectively~\cite{kaluva20182d}. Bi et~al. integrated 2D residual blocks into their network and gained the Dice score of 0.959~\cite{bi2017automatic}. Moreover, Li et~al. built a hybrid densely connected U-Net for liver and tumor segmentation, which combined both 2D and 3D features on liver and tumor~\cite{Li2017H}. They reached Dice scores of 0.961 and 0.722 on liver and tumor segmentation respectively. Pandey et~al. reduced the complexity of deep neural network by introducing ResNet-blocks and obtained the Dice score of 0.587 on tumor segmentation~\cite{pandey2018segmentation}. However, as mentioned earlier, most of them segmented the liver or lesion regions based on 2D slices from 3D volumes. The spatial information has not been taken into account to the maximum extent.

Recently, attention based image classification~\cite{wang2017residual} and semantic segmentation architectures~\cite{chen2016attention} have attracted a lot of attentions. Some medical imaging tasks have been dealt with using the attention mechanism to solve the issues in real application. For instance, Schlemper et~al. proposed an attention-gated networks for real-time automated scan plane detection in fetal ultrasound screening~\cite{schlemper2018attention}. The integrated self-gated soft-attention mechanisms, which can be easily incorporated into other networks, achieved good performance. Overall, it is expected that 3D deep networks combined with the attention mechanism would achieve a good performance for liver/tumor extraction tasks.

\section{Methodology}
\label{sec:methodology}
\subsection{Overview of our proposed architecture}

Our overall architecture for segmentation is depicted in Fig.~\ref{fig:overview of architecture}. The proposed architecture consists of three main stages which extract liver and tumor sequentially. Firstly, in order to reduce the overall computational time, we used a 2D residual attention-aware U-Net (RA-UNet) named RA-UNet-\Rmnum{1} based on a residual attention mechanism and U-Net connections to mark out a coarse liver boundary box. Next, a 3D RA-UNet, which is called RA-UNet-\Rmnum{2}, was trained to obtain a precise liver VOI. Finally, the prior liver VOI was sent to a second RA-UNet-\Rmnum{2} to extract the tumor region. The designed network can handle volumes in various complicated conditions and obtain desirable results in different liver/tumor datasets.

\subsection{Datasets and materials}

In our study, we used the public Liver Tumor Segmentation Challenge (LiTS) dataset to evaluate the proposed architecture. It has a total of 200 CT scans containing 130 scans as training data and 70 scans as test data, both of which have the same 512$\times$512 in-plane resolution but with different number of axial slices in each scan. These training data and their corresponding ground truth are provided by various clinical sites around the world, while the ground truth of the test data is not available.

Another dataset named 3DIRCADb is used as an external test dataset to test the generalization and scalability of our model. It includes 20 enhanced CT scans and the corresponding manually segmented tumors from European hospitals. The number of axial slices, which have 512$\times$512 in-plane resolution, differs for each scan. 

\subsection{Data preprocessing}

For a medical image volume, Hounsfield units (HU) is a measurement of relative densities determined by CT. Normally, the HU values range from -1000 to 1000. Because tumors grow on the liver tissue, the surrounding bones, air, or irrelevant tissues may disturb the segmentation result. Hence, an initial segmentation was used to filter out those noises, leaving the liver region clean to be segmented. In terms of convenience and efficiency, we took a global windowing step as our data preprocessing strategy.

We list the typical radiodensities of some main tissues in Table~\ref{table:Typical radiodensities}, which shows that these tissues have a wide range of HU values. From the table, the HU value for air is typically above -200; for bone it is the highest HU values among these tissues; for liver it is from 40 HU to 50 HU; for water it is approximately from -10 HU to 10 HU; and for blood it is from 3 HU to 14 HU.

\begin{table}
\caption{Typical tissues radiodensities of human body}
\begin{center}
\renewcommand\tabcolsep{30.5pt}    
\begin{tabular}{cc}
\toprule
Tissue & HU \\
\midrule
Air & -200+ \\
Bone & 400+ \\
Liver & 40$\sim$50\\
Water & 0$\pm$10\\
Blood & 3$\sim$14 \\
\bottomrule
\end{tabular}
\end{center}
\label{table:Typical radiodensities}
\end{table}

\begin{figure}
\centering
\includegraphics[scale=0.4]{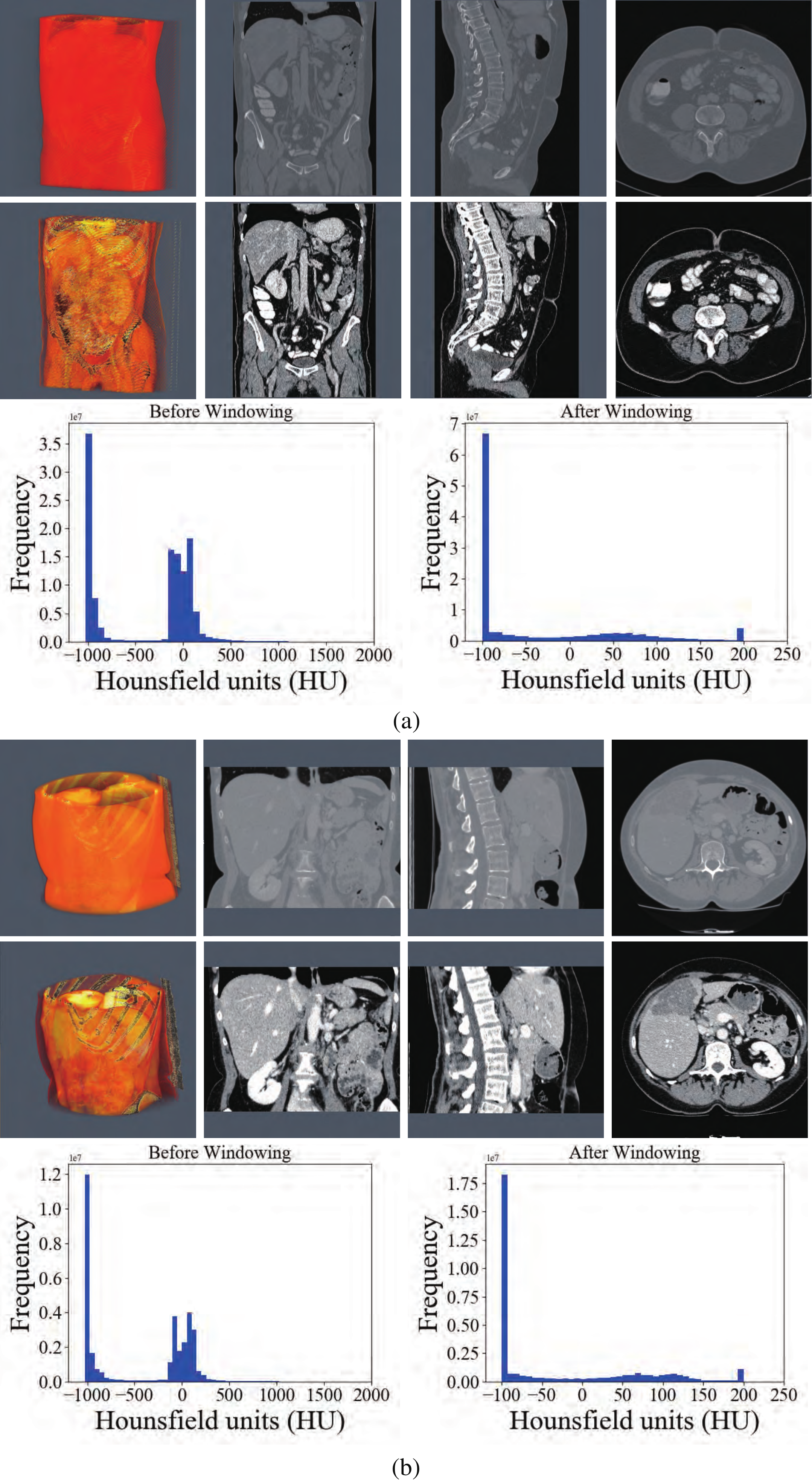}
\caption{Comparison between the raw CT scans (first row), windowed (second row) scans and histograms of HU (third row) before and after windowing. (a) shows the comparison on LiTS. (b) shows the comparison on 3DIRCADb.}
\label{fig:overview of preprocessing}
\end{figure}

In this article, we set the HU window at the range from -100 to 200. With such a window, irrelevant organ and tissues were mostly removed. The first rows of Fig.~\ref{fig:overview of preprocessing}(a) and (b) show the 3D, coronal, sagittal, and axial plane views of the raw volumes of LiTS and 3DIRCADb respectively. The second rows show the preprocessed volumes with irrelevant organ removed. It can be seen that most of the noise has been removed. The distribution of HU values before and after windowing is illustrated on the left and right of the third rows in Fig.~\ref{fig:overview of preprocessing}(a) and (b) where Frequency denotes the frequency of HU values. We applied the zero-mean normalization and min-max normalization on the data after the windowing. No more image processing was performed.

\begin{figure*}[ht]
\centering
\includegraphics[scale=0.6]{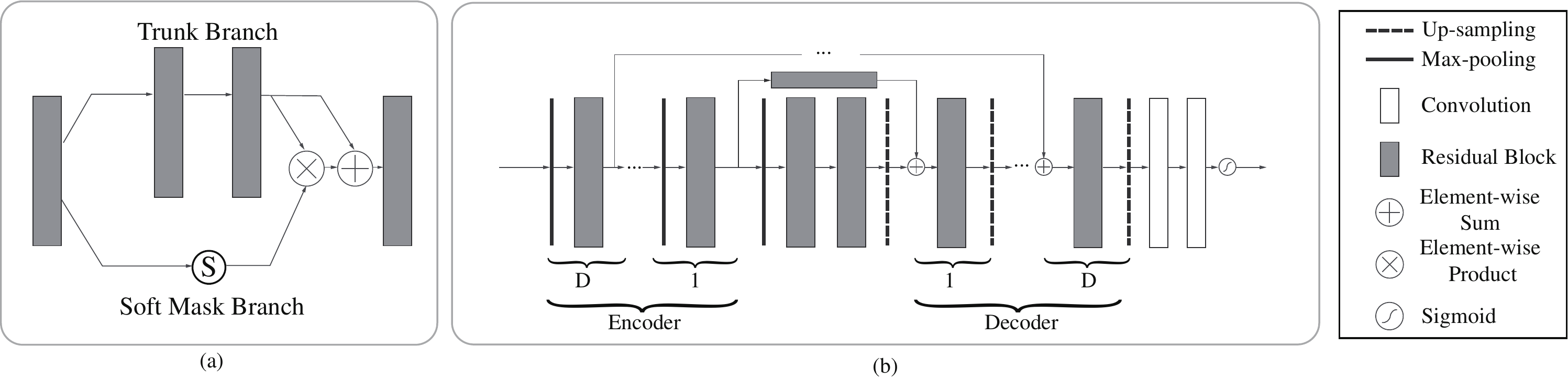}
\caption{The architecture of the attention residual module. (a) The attention residual module contains a trunk branch and a soft mask branch. The trunk branch learns original features while the soft mask branch focuses on reducing noises and enhancing good features. (b) The soft mask branch contains a stack of encoder-decoder blocks. D denotes the depth of skip connections. In our network, we set D to 0,1,2,3 according to the specific location of the attention residual block.}
\label{fig:overview of attention residual}
\end{figure*}

\begin{figure}
\centering
\includegraphics[scale=0.9]{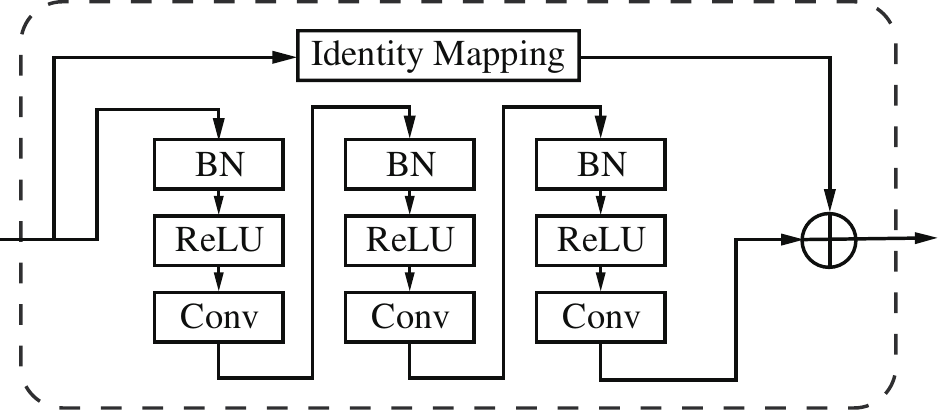}
\caption{Sample of a residual block in the dashed window. An identity mapping and convolutional blocks are added before the final feature output.}
\label{fig:overview of residual}
\end{figure}

\subsection{RA-UNet architecture}
The first time that an attention mechanism was introduced in semantic image segmentation was in~\cite{chen2016attention}, which combined $\emph{share-net}$ with attention mechanisms and achieved good performance. More recently, the attention mechanism is gradually applied to medical image segmentation~\cite{oktay2018attention},~\cite{schlemper2018attention}. Inspired by residual attention learning~\cite{wang2017residual} and U-Net~\cite{ronneberger2015u}, we propose the RA-UNet that has a ``very deep" architecture for the liver tumor segmentation task. The residual block allows a network to have hundreds of layers, while the attention mechanism learns to focus on locations that are relevant for discriminating object of interest. The overview of the architecture is depicted in Fig.~\ref{fig:overview of architecture}(d).

\subsubsection{U-Net as the basic architecture}
Our RA-UNet has an overall architecture similar to the standard U-Net, consisting of an encoder and a decoder symmetrically on the two sides of the architecture. The contextual information is propagated by the encoder within the rich skip connections which enables the extraction of hierarchical features with more complexity. The decoder receives features that have diverse complexity and reconstructs the features in a coarse-to-fine manner. A notable innovation is that the U-Net introduces long-range connections through the encoder part and the corresponding decoder part, so that different hierarchical features from the encoder can be merged to the decoder which makes the network much more precise and expansible.

\subsubsection{Residual learning mechanism}
The network depth is of crucial importance. However, gradient vanishing is a common problem in a very deep neural network when carrying out back propagation, which results in poor training results. In order to overcome this problem, He et~al. proposed the deep residual learning framework to learn the residual of the identity map~\cite{he2016deep}. In our study, residual blocks are stacked except the first layer and the last layer (Fig.~\ref{fig:overview of architecture}(d)) to unleash the capability of deep neural network and make it go ``deeper". The stacked residual blocks solve the gradient vanishing problem at the structural level of the neural network by using identity mappings as the skip connections and after-addition activation. The residual units directly propagate features from early convolution to late convolution and improve the performance of the model consequently. The residual block is defined as:
\setlength{\abovedisplayshortskip}{0pt}
\setlength{\belowdisplayshortskip}{0pt}
\begin{equation}
\textbf{\emph{OR}}_{i,c}(\textbf{\emph{x}})=\textbf{\emph{x}}+\textbf{\emph{f}}_{i,c}(\textbf{\emph{x}})
\label{eq:Residual learning mechanism}
\end{equation}                                   
where $\textbf{\emph{x}}$ denotes the first input of a residual block,  $\textbf{\emph{OR}}$ denotes the output of a residual block, $i$ ranges over all spatial positions, $\emph{c}\in \{ 1,...,\emph{C} \}$ indicates the index of channels, $\emph{C}$ is the total number of channels, and $\textbf{\emph{f}}$ represents the residual mapping to be learned.
 
The residual block consists of three sets of combination of a batch normalization (BN) layer, an activation (ReLU) layer, and a convolutional layer. A convolutional identity mapping connection is used to ensure the accuracy as the network goes ``deeper"~\cite{he2016deep}. The detailed residual unit is illustrated in Fig.~\ref{fig:overview of residual}.

\subsubsection{Attention residual mechanism}
The performance will drop if only naive stacking is used for the attention modules. This can be solved by the attention residual learning proposed by Wang et~al.~\cite{wang2017residual}. The attention residual mechanism divides the attention module into a trunk branch and a soft mask branch, where the trunk branch is used to process the original features and the soft mask branch is used to construct the identity mapping. The output $\textbf{\emph{OA}}$ of the attention module under attention residual learning can be formulated as:
\setlength{\abovedisplayshortskip}{0pt}
\setlength{\belowdisplayshortskip}{0pt}
\begin{equation}
\textbf{\emph{OA}}_{i,c}(\textbf{\emph{x}})=(1+\textbf{\emph{S}}_{i,c}(\textbf{\emph{x}}))\textbf{\emph{F}}_{i,c}(\textbf{\emph{x}})
\label{eq:Attention residual mechanism}
\end{equation}
where $\textbf{\emph{S}}(\textbf{\emph{x}})$ has values in [0,1]. If $\textbf{\emph{S}}(\textbf{\emph{x}})$ is close to 0, $\textbf{\emph{OA}}(\textbf{\emph{x}})$ will approximate the original feature maps $\textbf{\emph{F}}(\textbf{\emph{x}})$. The soft mask branch $\textbf{\emph{S}}(\textbf{\emph{x}})$, which selects identical features and suppress noised from the trunk branch, plays the most important role in the attention residual mechanism.

The soft mask branch has an encoder-decoder structure which has been widely applied to medical image segmentation~\cite{cciccek20163d},~\cite{ronneberger2015u},~\cite{alom2018recurrent}. In the attention residual mechanism, it is designed to enhance good features and reduce the noises from the trunk branch. The encoder in the soft mask branch contains a max-pooling operation, a residual block, and a long-range residual block connected to the corresponding decoder, where an element-wise sum is performed following a residual block and an up-sampling operation. After the encoder and decoder parts of the soft mask, two convolutional layers and one sigmoid layer are added to normalize the output. Fig.~\ref{fig:overview of attention residual} illustrates the attention residual module in details.

In general, the attention residual mechanism can keep the original feature information through the trunk branch and pay attention to those liver tumor features by the soft mask branch. By using the attention residual mechanism, our RA-UNet can improve the performance significantly.

\subsubsection{Loss function}
The weights are learnt by minimizing the loss function. We employed a loss function based on the Dice coefficient proposed in~\cite{milletari2016v-net:} in this study. The loss $\textbf{\emph{L}}$ is defined as follows:
\setlength{\abovedisplayshortskip}{0pt}
\setlength{\belowdisplayshortskip}{0pt}
\begin{equation}
\textbf{\emph{L}}=1-\frac{2 \sum_{i=1}^{N}\textbf{\emph{s}}_{i}\textbf{\emph{g}}_{i}}{\sum_{i=1}^{N}\textbf{\emph{s}}_{i}^2+\sum_{i=1}^{N}\textbf{\emph{g}}_{i}^2}
\end{equation}
where $N$ is the number of voxels, $\textbf{\emph{s}}_i$ and $\textbf{\emph{g}}_i$ belong to the binary segmentation and binary ground truth voxel sets respectively. The loss function measures the similarity of two samples directly.

\subsection{Liver localization using RA-UNet-\Rmnum{1}}

The first stage aimed to locate the 3D liver boundary box. A 2D version RA-UNet-\Rmnum{1} was introduced here to segment a coarse liver region, which can reduce the computational cost of the subsequent RA-UNet-\Rmnum{2}, remove the redundant information, and provide more effective information. It worked as a ``baseline" to limit the scope of the liver. Table~\ref{table:RA-UNet-1 params} illustrates the detailed network parameters. The network went down from the top to the bottom in the encoder, and reversed in the decoder. During the encoding phase, the RA-UNet-\Rmnum{1} received a single-channel and down sampled the 256$\times$256-sized slices and passed them down to the bottom. During the decoding phase, lower features were passed from the bottom to the top with resolution doubled through the up-sampling operation. Note that the long-range connection between the encoder and the decoder was realised by the attention block. We then combined the features from the attention blocks with those from the corresponding up-sampling level in the decoder via concatenation. Then the concatenated features were passed on to the decoder. Finally, a convolutional layer with a 3$\times$3 kernel size was used to generate the final probability map of liver segmentation.

During the testing phase, we down sampled the slices to 256$\times$256 and fed the preprocessed slices into the trained RA-UNet-\Rmnum{1} model. Next, we stacked all the slices in their original sequence. Then a 3D connect-component labeling~\cite{hossam20103d} was employed, and the largest component was chosen as the coarse liver region. Finally, we interpolated the liver region to its original volume size with a 512$\times$512 in-plane resolution.

\begin{table}[]
\caption{Architecture of the proposed RA-UNet-\Rmnum{1}. Here [ ] denotes the long range connection; [ , ] denotes the concatenate operation; Conv means the convolution; Up stands for the up-sampling; Res denotes the residual block; and Att denotes the attention block}
\renewcommand\tabcolsep{6.5pt}        
\begin{tabular}{|c|c|c|c|c|}
\hline
Encoder & Output size               & Decoder &                     & Output size               \\ \hline
Input   & 256\textasciicircum{}2$\times$1  & Att1    & {[}Res4{]}, depth=0 & 16\textasciicircum{}2$\times$128 \\ \hline
Conv1   & 256\textasciicircum{}2$\times$16 & Res7    & {[}Up1, Att1{]}     & 16\textasciicircum{}2$\times$128 \\ \hline
Pooling & 128\textasciicircum{}2$\times$16 & Up2     &                     & 32\textasciicircum{}2$\times$128 \\ \hline
Res1    & 128\textasciicircum{}2$\times$16 & Att2    & {[}Res3{]}, depth=1 & 32\textasciicircum{}2$\times$64  \\ \hline
Pooling & 64\textasciicircum{}2$\times$16  & Res8    & {[}Up2, Att2{]}     & 32\textasciicircum{}2$\times$64  \\ \hline
Res2    & 64\textasciicircum{}2$\times$32  & Up3     &                     & 64\textasciicircum{}2$\times$64  \\ \hline
Pooling & 32\textasciicircum{}2$\times$32  & Att3    & {[}Res2{]}, depth=2 & 64\textasciicircum{}2$\times$32  \\ \hline
Res3    & 32\textasciicircum{}2$\times$64  & Res9    & {[}Up3, Att3{]}     & 64\textasciicircum{}2$\times$32  \\ \hline
Pooling & 16\textasciicircum{}2$\times$64  & Up4     &                     & 128\textasciicircum{}2$\times$32 \\ \hline
Res4    & 16\textasciicircum{}2$\times$128 & Att4    & {[}Res1{]}, depth=3 & 128\textasciicircum{}2$\times$16 \\ \hline
Pooling & 8\textasciicircum{}2$\times$128  & Res10   & {[}Up4, Att4{]}     & 128\textasciicircum{}2$\times$16 \\ \hline
Res5    & 8\textasciicircum{}2$\times$256  & Up5     &                     & 256\textasciicircum{}2$\times$16 \\ \hline
Res6   & 8\textasciicircum{}2$\times$256  & Conv2   & {[}Up5, Conv1{]}    & 256\textasciicircum{}2$\times$16 \\ \hline
Up1     & 16\textasciicircum{}2$\times$256 & Conv3   &                     & 256\textasciicircum{}2$\times$1  \\ \hline
\end{tabular}
\label{table:RA-UNet-1 params}
\end{table}

\begin{table}[]
\caption{Architecture of the proposed RA-UNET-\Rmnum{2}. Here [ ] denotes the long range connection; [ , ] denotes the concatenate operation; Conv means the convolution; Up stands for the up-sampling; Res denotes the residual block; and Att denotes the attention block}
\renewcommand\tabcolsep{4.5pt}      
\begin{tabular}{|c|c|c|c|c|}
\hline
Encoder & Output size                  & Decoder &                     & Output size                  \\ \hline
Input   & 224\textasciicircum{}2$\times$32$\times$1  & Att1    & {[}Res4{]}, depth=0 & 14\textasciicircum{}2$\times$2$\times$256  \\ \hline
Conv1   & 224\textasciicircum{}2$\times$32$\times$32 & Res7    & {[}Up1, Att1{]}     & 14\textasciicircum{}2$\times$2$\times$256  \\ \hline
Pooling & 112\textasciicircum{}2$\times$16$\times$32 & Up2     &                     & 28\textasciicircum{}2$\times$4$\times$256  \\ \hline
Res1    & 112\textasciicircum{}2$\times$16$\times$32 & Att2    & {[}Res3{]}, depth=1 & 28\textasciicircum{}2$\times$4$\times$128  \\ \hline
Pooling & 56\textasciicircum{}2$\times$8$\times$32   & Res8    & {[}Up2, Att2{]}     & 28\textasciicircum{}2$\times$4$\times$128  \\ \hline
Res2    & 56\textasciicircum{}2$\times$8$\times$64   & Up3     &                     & 56\textasciicircum{}2$\times$8$\times$128  \\ \hline
Pooling & 28\textasciicircum{}2$\times$4$\times$64   & Att3    & {[}Res2{]}, depth=2 & 56\textasciicircum{}2$\times$8$\times$64   \\ \hline
Res3    & 28\textasciicircum{}2$\times$4$\times$128  & Res9    & {[}Up3, Att3{]}     & 56\textasciicircum{}2$\times$8$\times$64   \\ \hline
Pooling & 14\textasciicircum{}2$\times$2$\times$128  & Up4     &                     & 112\textasciicircum{}2$\times$16$\times$64 \\ \hline
Res4    & 14\textasciicircum{}2$\times$2$\times$256  & Att4    & {[}Res1{]}, depth=3 & 112\textasciicircum{}2$\times$16$\times$32 \\ \hline
Pooling & 7\textasciicircum{}2$\times$1$\times$256   & Res10   & {[}Up4, Att4{]}     & 112\textasciicircum{}2$\times$16$\times$32 \\ \hline
Res5    & 7\textasciicircum{}2$\times$1$\times$512   & Up5     &                     & 224\textasciicircum{}2$\times$32$\times$32 \\ \hline
Res6    & 7\textasciicircum{}2$\times$1$\times$512   & Conv2   & {[}Up5, Conv1{]}    & 224\textasciicircum{}2$\times$32$\times$32 \\ \hline
Up1     & 14\textasciicircum{}2$\times$2$\times$512  & Conv3   &                     & 224\textasciicircum{}2$\times$32$\times$1  \\ \hline
\end{tabular}
\label{table:RA-UNet-2 params}
\end{table}

\subsection{Liver segmentation using RA-UNet-\Rmnum{2}}

The RA-UNet-\Rmnum{2} was a 3D model which fully utilized the volume information and captured the spatial information. The 3D U-Net type architecture~\cite{cciccek20163d} would merge the low resolution and high resolution features to generate an accurate segmentation. Meanwhile, the residual blocks would handle the gradient vanishing problem, allowing the network to go ``deeper" without accuracy degradation. In addition, using large image patches (224$\times$224$\times$32) for training provides much richer contextual information than using small image patches, and this usually leads to more global segmentation results. The RA-UNet-\Rmnum{2} has less parameters than the traditional U-Net~\cite{ronneberger2015u}. With this architecture, the number of parameters has been largely decreased to only 4M training parameters while reaching the depth of 641. During the training phase, we interpolated the liver boundary box in the $x{-}y$ plane to a fixed size and randomly picked a number of 32 slices successively in the $z$ direction to form the training patches for RA-UNet-\Rmnum{2}.

During the testing phase, RA-UNet-\Rmnum{2} was employed on each CT patch to generate 3D liver probability patches in sequence. Then, we interpolated and stacked those probability patches to be restored to the original size of the boundary box. A voting strategy was used to generate the final liver probability of VOI from overlapped sub-patches. A 3D connect-component labeling was used and the largest component was chosen on the merged VOI to yield the final liver region. Detailed network parameters were listed in Table~\ref{table:RA-UNet-2 params}. The network received 224$\times$224$\times$32 patches and generated the output for the probability volume of patches.

\subsection{Extraction of tumors based on RA-UNet-\Rmnum{2}}

\begin{figure}
\centering
\includegraphics[scale=0.9]{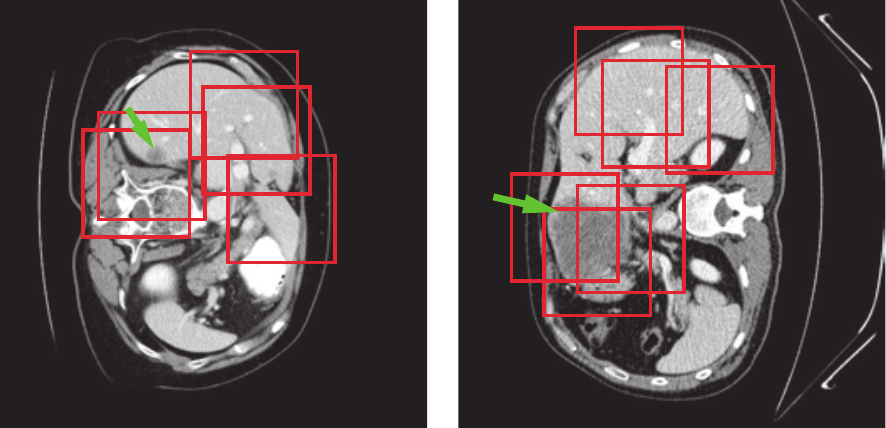}
\caption{Tumor patch extraction results. The green arrows point to the tumor regions and the red boxes show the patches used for training.}
\label{fig:overview of tumor extraction}
\end{figure}

Tumor region extraction was similar to liver segmentation but no interpolation and resizing were performed. Because the size of the tumor is much smaller than that of the liver, original tumor resolution was used to avoid losing small lesions. Furthermore, in order to solve the data imbalance issue and learn more effective tumor features, we picked patches on both tumor and its surroundings non-tumor regions for training as shown in Fig.~\ref{fig:overview of tumor extraction}. Note that only those in the liver VOIs would be the candidate patches for training. 

During the testing phase, we extracted the tumors following a similar routine as for the liver segmentation step except the use of interpolation. Subsequently, a voting strategy is used again on the merged VOI to yield the final tumor segmentation. At last, we filtered out those voxels which were not in the liver region.

\begin{figure}
\centering
\includegraphics[scale=0.5]{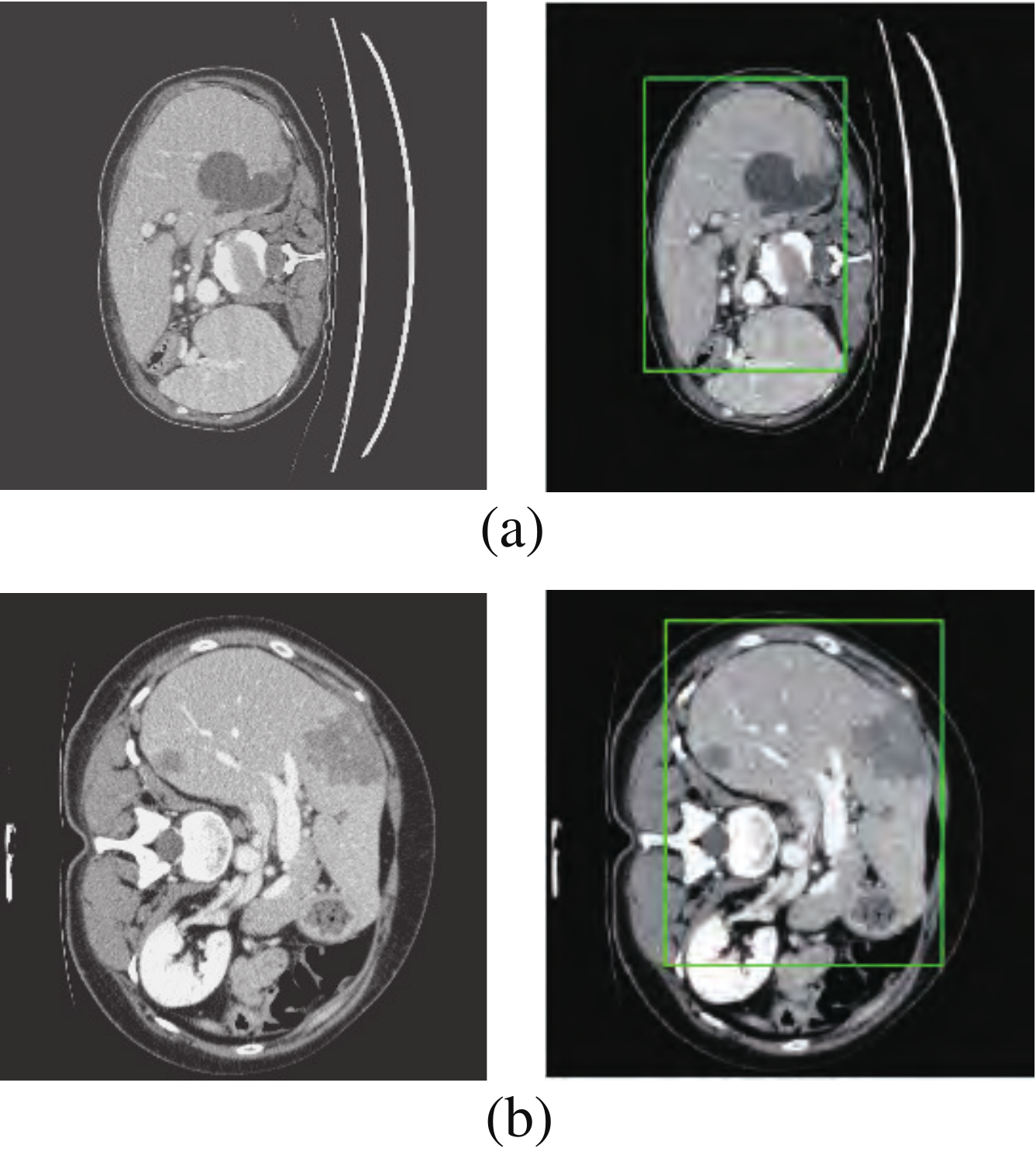}
\caption{Liver localization using RA-UNet-\Rmnum{1}. From left to right the figure shows the preprocessed slice, and the final boundary box which restricts the liver region. (a) A typical slice from the LiTS validation dataset. (b) A typical slice from the 3DIRCADb dataset. The RA-UNet-\Rmnum{1} enables the coarse localization of liver regions.}
\label{fig:overview of liver localization}
\end{figure}

\subsection{Evaluation metrics}

We evaluated the performance of the proposed approach using the metrics introduced in~\cite{heimann2009comparison}. The evaluation metrics include Dice score (DS)~\cite{wu2016automatic}, which confounds both detection and segmentation, consist of Dice global (Dice score computed on all combined volumes denoted with DG) and Dice per case (mean Dice score per volume denoted with DC), Jaccard similarity coefficient (Jaccard), volumetric overlap error (VOE), relative volume difference (RVD), average symmetric surface distance (ASSD), and maximum surface distance (MSD).

\subsection{Implementation details}

The RA-UNet architecture was constructed using the Keras~\cite{chollet_keras_2015} and the TensorFlow~\cite{abadi2015tensorflow} libraries. All the models were trained from scratch. The parameters of the network were initialized with random values and then they were trained with back-propagation based on Adam~\cite{kingma2014adam} with an initial learning rate (LR) of 0.001, $\beta_1$=0.9, and $\beta_2$=0.999. The learning rate would be reduced to LR$\times$0.1 if the network went to plateau after 20 epoches. We used 5-fold cross training on the LiTS training dataset, and evaluated the performance on the LiTS test dataset. To demonstrate the generalization of our RA-UNet, we also evaluated the performance on the 3DIRCADb dataset using the well-trained weights from the LiTS training dataset. For the liver and tumor trainings, the total numbers of epoches were set at 50 and 50 for each fold respectively. An integration operation by a voting strategy is implemented to ensemble all the prediction results of 5 models. The training of all the models was performed with an NVIDIA 1080Ti GPU.

\section{Experiments and results}
\label{sec:exp_and_res}
\subsection{Liver volume of interest localization}

We first down sampled the input slices to a 256$\times$256 in plane resolution to simplify computation. In order to reduce the computation cost, we used all the slices which have liver on the images together with 1/3 of those randomly picked slices without liver as the training data. There are a total of 32,746 slices with liver which were used, including 23,283 slices for training and 9,463 slices for validation. Note that 5-fold training was not employed at this stage, because our goal at this stage was to obtain a coarse liver boundary box and reduce the computational time.

After stacking all the slices and employing the 3D connect-component labeling, we calculated the 3D boundary box of the slices with liver, and extended 10 pixels in coronal, sagittal, and axial directions to ensure that the entire liver region was included. Fig.~\ref{fig:overview of liver localization} shows the liver localization results from RA-UNet-\Rmnum{1}. It demonstrates that the attention mechanism has successfully constrained the liver region, and RA-UNet-\Rmnum{1} can greatly restrict the liver region within a boundary box.

\subsection{Liver segmentation using RA-UNet-\Rmnum{2}}

RA-UNet-\Rmnum{2} allowed the network to go ``deeper". However, the implementation of a 3D network is limited by the hardware and memory requirements~\cite{prasoon2013deep}. In order to balance the computational cost and efficiency, we first carried out interpolation in the region inside the liver boundary box to the size of 224$\times$224$\times \emph{M}$, where $\emph{M}$ was the axial length of the liver boundary box. Then we cropped the volumetric patches (224$\times$224$\times$32) randomly from each boundary box, which was constrained by the liver boundary box. Totally, 5,096 patches were selected for training and validation.

\begin{figure}
\centering
\includegraphics[scale=0.45]{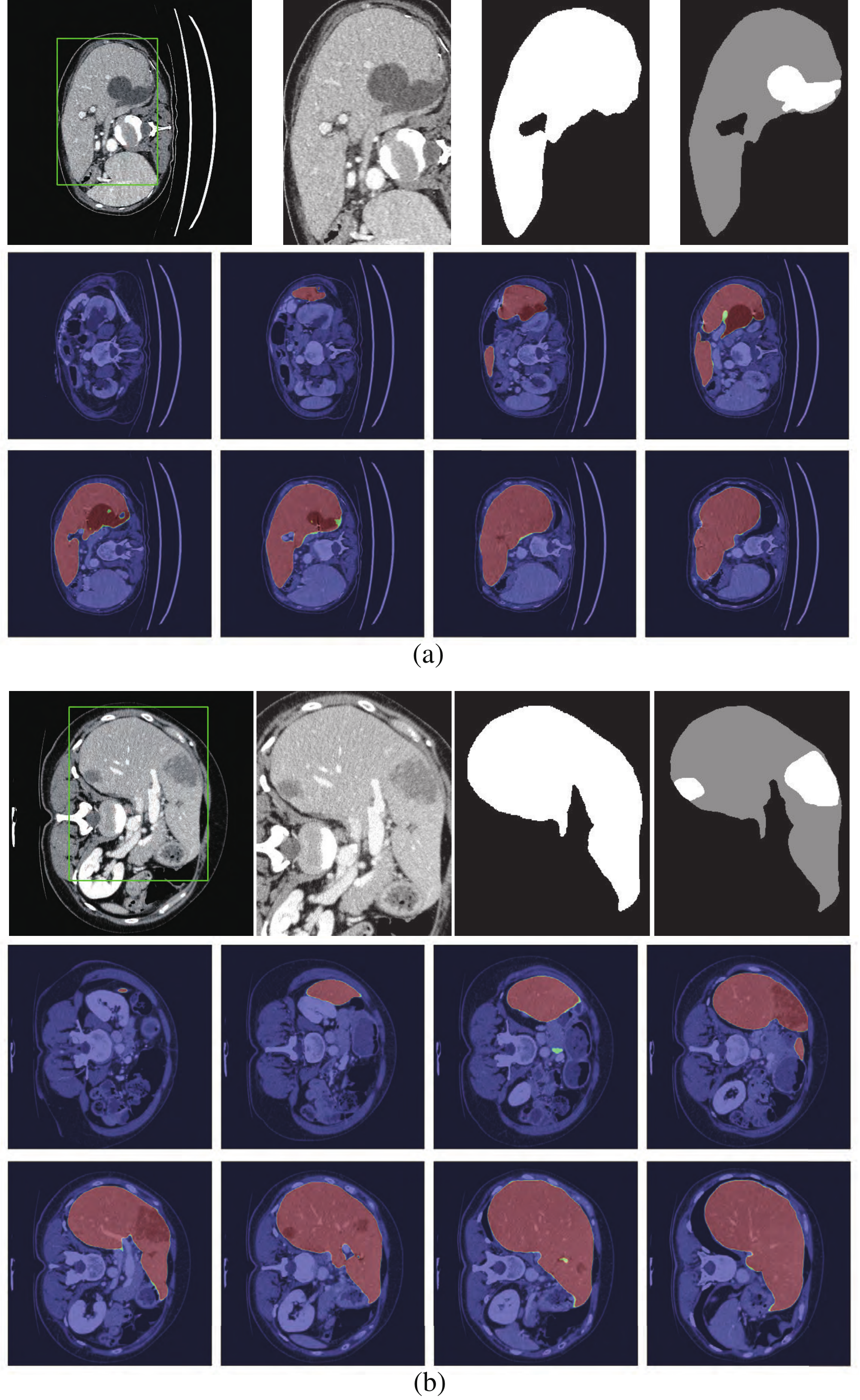}
\caption{Liver segmentation results based on RA-UNet-\Rmnum{2}. (a) is from the LiTS validation dataset and (b) is from the 3DIRCADb dataset. From left to right, the first row of each subplot shows the liver in the green boundary box, magnified liver region, the liver segmentation results, and the corresponding ground truth. The second and the third rows show the probability heat map of liver segmentation results. The darker the color is, the higher the possibility of the liver region is. Note that the ground truth contains liver in gray and tumor in white.}
\label{fig:overview of liver segmentation}
\end{figure}

Fig.~\ref{fig:overview of liver segmentation} shows the liver segmentation based on RA-UNet-\Rmnum{2}, which indicates that our proposed network has the ability to learn 3D contextual information and could successfully extract the liver from adjacent slices in an image volume. After the 3D connect-component labeling was carried out, the liver region was precisely extracted by selecting the largest region.

As shown in Table~\ref{table:liver segmentation metrics}, our method reached up to 0.961 and 0.977 Dice score on the LiTS test dataset and the 3DIRCADb dataset respectively. It reveals that RA-UNet yields remarkable liver segmentation results. Then we can extract tumors from the segmented liver regions.

\begin{table}[]
\caption{Scores of the liver segmentation on the LiTS test dataset and the 3DIRCADb dataset}
\begin{center}
\renewcommand\tabcolsep{25.5pt}        
\begin{tabular}{ccc}
\toprule
& LiTS & 3DIRCADb \\
\midrule
DC & 0.961 & 0.977 \\
Jaccard & 0.926 & 0.977 \\
VOE & 0.074 & 0.045 \\
RVD & 0.002 & -0.001 \\
ASSD & 1.214 & 0.587  \\
MSD & 26.948 & 18.617 \\
\bottomrule
\end{tabular}
\end{center}
\label{table:liver segmentation metrics}
\end{table}

\begin{figure}
\centering
\includegraphics[scale=0.6]{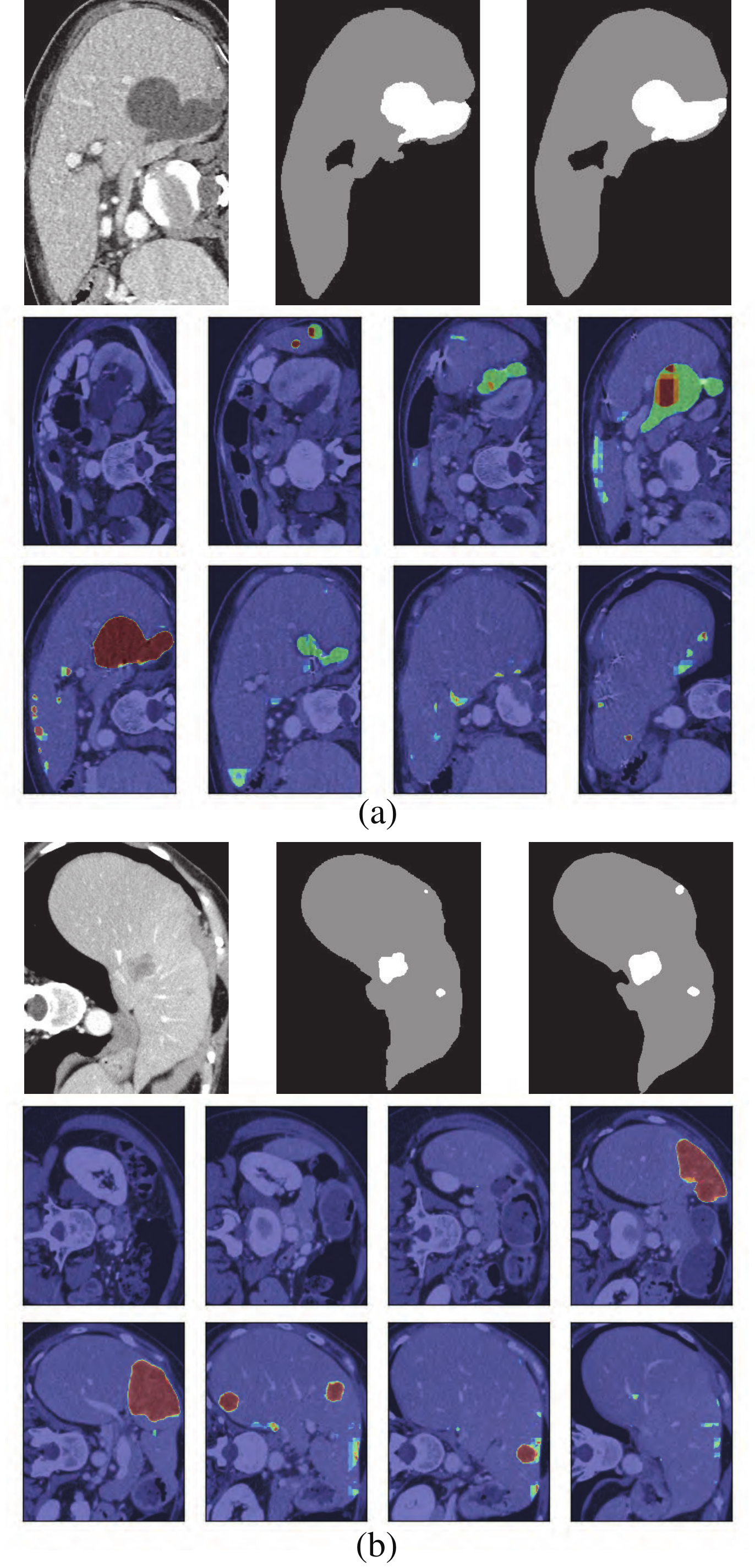}
\caption{Tumor segmentation results based on RA-UNet-\Rmnum{2}. (a) is from the LiTS validation dataset, and (b) is from the 3DIRCADb dataset. From left to right, the first row of each subplots indicates the raw images, segmentation results of liver tumor, and the corresponding ground truth. The second and the third rows show the probability heat map of tumor segmentation results.}
\label{fig:overview of tumor segmentation}
\end{figure}

\subsection{Extraction of tumors based on RA-UNet-\Rmnum{2}}

Tumors were tiny structures compared to livers. Therefore, no interpolation or resizing was applied on tumor patch sampling to avoid information loss from image scaling. It was difficult to decide what size of patch for training could reach a desirable performance. In order to determine the patch size, we set the patch size of 32$\times$32$\times$32, 64$\times$64$\times$32, and 128$\times$128$\times$32 respectively to test the performance of tumor segmentation. Results showed that 128$\times$128$\times$32 patch-sized data achieved a best tumor segmentation performance. The larger the patch size was, the richer context in formation the patches could provide. While due to the limitation of computational resource, 128$\times$128$\times$32 was chosen empirically for tumor patches. We randomly picked 150 patches from each liver volume in the boundary box. Totally, 17,700 patches were chosen from LiTS as training and validation datasets. As shown in Table~\ref{table:tumor segmentation metrics}, our method reached 0.595 and 0.830 Dice scores on the LiTS test dataset and the 3DIRCADb dataset respectively. Fig.~\ref{fig:overview of tumor segmentation} shows the tumor segmentation results in details.

\begin{figure}
\centering
\includegraphics[scale=0.25]{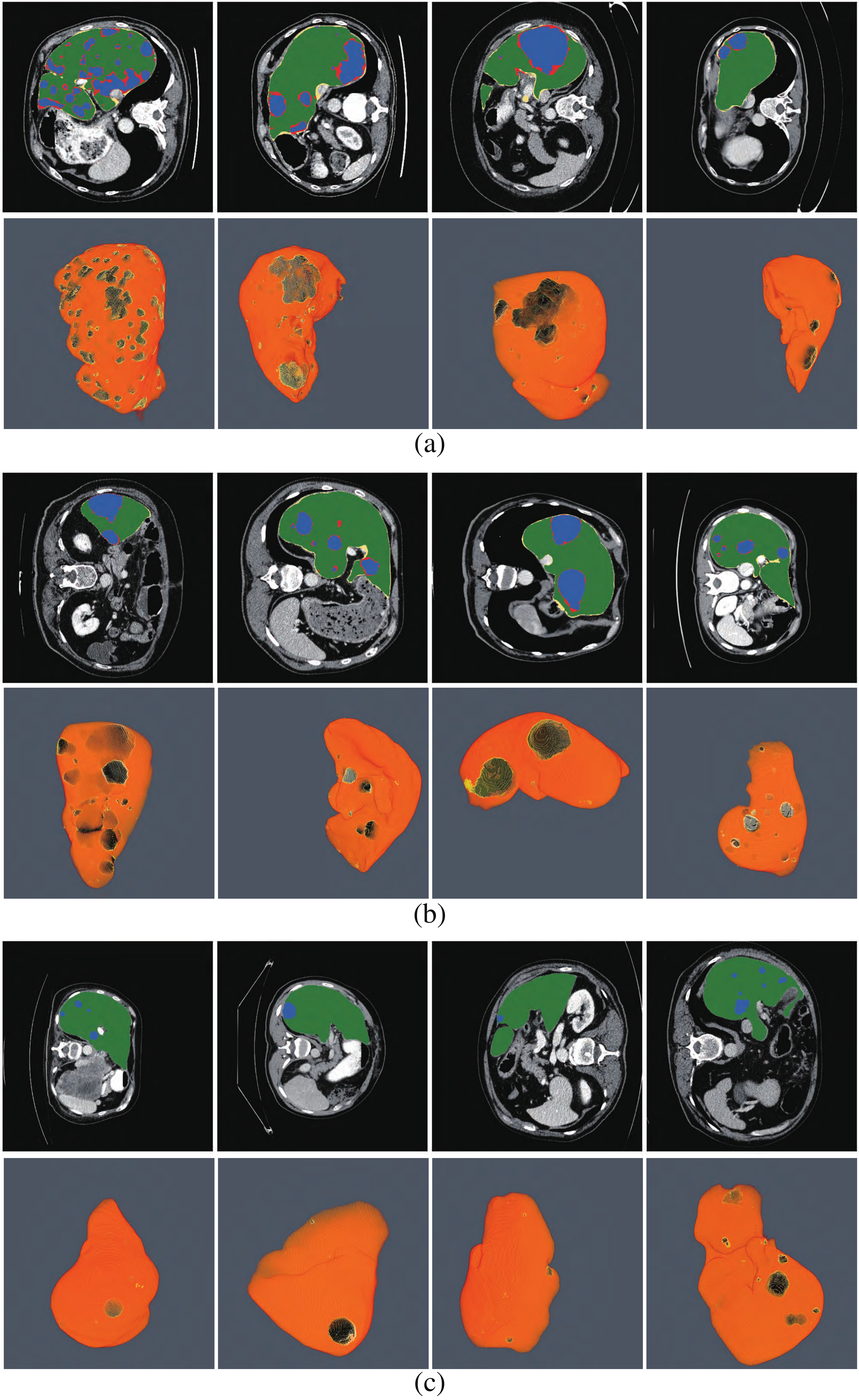}
\caption{Automatic liver and tumor segmentation with RA-UNet. The green regions indicate the correctly extracted liver, the yellow regions are the wrongly extracted liver, the blue color depicts the correctly extracted tumor regions and the red color means wrongly extracted tumor. The first row of each subplot shows four slices from different volumes in the axial view and the second row of each subplot shows the corresponding 3D view of the entire liver/tumor segmentation results. (a) is from the LiTS dataset. (b) is from the 3DIRCADb dataset. (c) shows the segmentation results on the LiTS test datasets. Note that no ground truth is provided for the LiTS test dataset.}
\label{fig:overview of final segmentation}
\end{figure}

Fig.~\ref{fig:overview of final segmentation} shows the liver/tumor segmentation results. It shows that liver regions which are large in size are successfully segmented and tumors that are tiny and hard to detect can be identified by the proposed method as well. Due to the low contrast with the surrounding livers and the extremely small size of some tumors, the proposed method still has some false positives and false negatives for tumor extraction.

\begin{table}[]
\caption{Scores of the tumor segmentation on the LiTS test dataset and the 3DIRCADb dataset}
\renewcommand\tabcolsep{25.5pt}        
\begin{center}
\begin{tabular}{ccc}
\toprule
& LiTS & 3DIRCADb \\
\midrule
DC & 0.595 & 0.830 \\
Jaccard & 0.611 & 0.744 \\
VOE & 0.389 & 0.255 \\
RVD & -0.152 & 0.740 \\
ASSD & 1.289 & 2.230  \\
MSD & 6.775 & 53.324 \\
\bottomrule
\end{tabular}
\end{center}
\label{table:tumor segmentation metrics}
\end{table}

\begin{table*}[]
\caption{Segmentation results compared with other methods on the LiTS test dataset}
\begin{center}
\renewcommand\tabcolsep{4.3pt} 
\renewcommand\arraystretch{1.3}      
\begin{tabular}{|c|c|c|c|c|c|c|c|c|c|c|c|c|c|c|c|}
\hline
\multirow{2}{*}{}      &           & \multicolumn{7}{c|}{LiTS Liver}                            & \multicolumn{7}{c|}{LiTS Tumor}                          \\ \cline{2-16} 
                       & Dimension & DC    & DG    & Jaccard & VOE   & RVD    & ASSD  & MSD     & DC    & DG    & Jaccard & VOE   & RVD    & ASSD  & MSD   \\ \hline
Kaluva et~al.~\cite{kaluva20182d} & 2D        & 0.912 & 0.923 & 0.850   & 0.150 & -0.008 & 6.465 & 45.928  & 0.492 & 0.625 & 0.589   & 0.411 & 19.705 & 1.441 & 7.515 \\ \hline
Bi et~al.~\cite{bi2017automatic}     & 2D        & 0.959 & -     & 0.922   & -     & -      & -     & -       & 0.500 & -     & 0.388   & -     & -      & -     & -     \\ \hline
Li et~al.~\cite{Li2017H}     & 2.5D      & 0.961 & 0.965 & -       & 0.074 & -0.018 & 1.450 & 27.118  & 0.722 & 0.824 & -       & 0.366 & 4.272  & 1.102 & 6.228 \\ \hline
MEDDIIR                & unknown   & 0.950 & 0.955 & -       & 0.094 & 0.047  & 1.597 & 28.911  & 0.658 & 0.819 & -       & 0.380 & -0.129 & 1.113 & 6.323 \\ \hline
Yuan~\cite{yuan2017hierarchical}          & 2D        & 0.963 & 0.967 & -       & 0.071 & -0.010 & 1.104 & 23.847  & 0.657 & 0.820 & -       & 0.378 & 0.288  & 1.151 & 6.269 \\ \hline
Summer                 & unknown   & 0.941 & 0.945 & -       & 0.108 & -0.066 & 6.552 & 152.350 & 0.631 & 0.786 & -       & 0.400 & -0.181 & 1.184 & 6.367 \\ \hline
\textbf{Proposed method}        & \textbf{3D}        & \textbf{0.961} & \textbf{0.963} & \textbf{0.926}   & \textbf{0.074} & \textbf{0.002}  & \textbf{1.214} & \textbf{26.948}  & \textbf{0.595} & \textbf{0.795} & \textbf{0.611}   & \textbf{0.389} & \textbf{-0.152} & \textbf{1.289} & \textbf{6.775} \\ \hline
\end{tabular}
\end{center}
\label{table:lits dataset score}
\end{table*}

\begin{table*}[]
\caption{Segmentation results compared with other methods on the 3DIRCADb dataset}
\begin{center}
\renewcommand\tabcolsep{12.9pt}  
\renewcommand\arraystretch{1.3}      
\begin{tabular}{|c|c|c|c|c|c|c|c|c|}
\hline
                      &           & \multicolumn{6}{c|}{3DIRCADb Liver}                & 3DIRCADb Tumor \\ \hline
                      & Dimension & DC    & Jaccard & VOE    & RVD    & ASSD  & MSD    & DC             \\ \hline
Chirst et~al.~\cite{christ2017automatic} & 2D        & 0.943 & -       & 0.107  & -0.014 & 1.6   & 24     & 0.56           \\ \hline
U-Net as in~\cite{ronneberger2015u}  & 2D        & 0.729 & -       & 0.39   & 0.87   & 19.4  & 119    & -              \\ \hline
Li et~al.~\cite{li2013likelihood}    & 2D        & 0.945 & -       & 0.068  & -0.112 & 1.6   & 28.2   & -              \\ \hline
Maya et~al.~\cite{eapen2015swarm}  & 3D        & -     & -       & 0.0554 & 0.0093 & 0.78  & 15.6   & -              \\ \hline
Lu et~al.~\cite{lu2017automatic}    & 3D        & -     & -       & 0.0936 & 0.0097 & 1.89  & 33.14  & -              \\ \hline
\textbf{Proposed method}       & \textbf{3D}        & \textbf{0.977} & \textbf{0.977}   & \textbf{0.045}  & \textbf{-0.001} & \textbf{0.587} & \textbf{18.617} & \textbf{0.83}           \\ \hline
\end{tabular}
\end{center}
\label{table:3dircadb dataset score}
\end{table*}

\subsection{Comparison with other methods}

There were several submissions about liver and tumor segmentation to the 2017 ISBI and MICCAI LiTS challenges. We reached a Dice per case of 0.961, Dice global of 0.963, Jaccard of 0.926, VOE of 0.074, RVD of 0.002, ASSD of 1.214, and MSD of 26.948, which is a desirable performance on the LiTS challenge for liver segmentation. For tumor burden evaluation, our method reached a Dice per case of 0.595, Dice global of 0.795, Jaccard of 0.611, VOE of 0.389, RVD of -0.152, ASSD of 1.289, and MSD of 6.775. Compared to other methods, Pandey et~al.'s~\cite{pandey2018segmentation} and Bellver et~al.'s~\cite{bellver2017detection} methods reached tumor Dice per case at 0.587 and 0.59 respectively, which were 2D segmentation methods. Our approach outperformed these two methods. The detailed results and all the performances are listed in Table~\ref{table:lits dataset score}. It is worth mentioning that our method was a full 3D segmentation technique with a much deeper network.

For the 3DIRCADb dataset, some works concentrated on liver segmentation, and there were a few about tumor segmentation. Hence, we listed the results of some approaches in Table~\ref{table:3dircadb dataset score}. Our methods reached a Dice per case of 0.977, Jaccard of 0.977, VOE of 0.045, RVD of -0.001, ASSD of 0.587, and MSD of 18.617, which show that our method performed significantly better than all the other methods on liver segmentation. It is worth mentioning that the proposed method was a 3D convolutional neural network and showed its generalization ability on the 3DIRCADb dataset using well-trained weights based on the LiTS dataset. Since most of the works aimed at liver segmentation, few of them displayed tumor segmentation results, we only compared with Chirst et~al.~\cite{christ2017automatic} on the 3DIRCADb dataset. It was worth mentioning that our method reached a mean Dice score of 0.830 on livers with tumors compared to a mean Dice score of 0.56 for the method in Chirst et~al.~\cite{christ2017automatic}.

\subsection{Extension to brain tumor segmentation}

Our 3D RA-UNet is extendable to other tumor segmentation tasks and shows its strong generalization ability. We used the Brain Tumor Segmentation Challenge (BraTS) 2018 dataset~\cite{menze2015multimodal},~\cite{bakas2017advancing} for validating our model. The BraTS2018 dataset contains 285 training data with 210 high-grade glioma (HGG) patients and 75 low-grade glioma (LGG) patients, and validation data with 66 patients. For each patient, the BraTS2018 training dataset provides 4 MRI 3D scans (T1, T1Gd, T2, and FLAIR) with a 155$\times$240$\times$240 resolution and the corresponding ground truth, while the validation data does not contain ground truth. The ground truth marks out background, necrosis (NCR), edema (ED), non-enhancing tumor (NET), and enhancing tumor (ET) with different labels. The labels in the provided data are: 1 for NCR \& NET, 2 for ED, 4 for ET, and 0 for no-tumor regions. To show the generalization capability of our RA-UNet, we also used the validation data of the Brain Tumor Segmentation Challenge (BraTS) 2017 dataset for testing using the well-trained weights from BraTS2018. The BraTS2017 dataset is similar to the BraTS2018 dataset, and more information can be found in~\cite{menze2015multimodal},~\cite{bakas2017advancing}.

\begin{figure}
\centering
\includegraphics[scale=0.4]{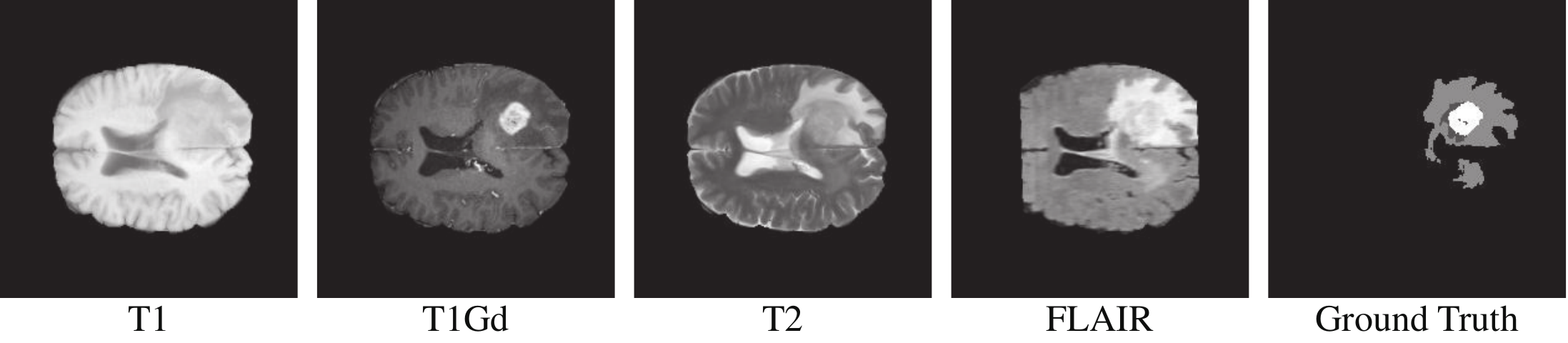}
\caption{Typical slices of a patient with four MRI scans and corresponding ground truth.}
\label{fig:overview of bratsdata}
\end{figure}

Fig.~\ref{fig:overview of bratsdata} shows a typical slide of a brain scan and its ground truth from the BraTS2018 dataset. According to~\cite{chen2017voxresnet}, organs could be robustly examined with multiple imaging modalities. We used single modality and multi-modality images to train RA-UNet sequentially, and it turned out that full tumor information could be provided by feeding multi-modality images. Thus, we concatenated all the modality data, and normalized them to [0,1]. No other preprocessing strategy was performed. This task aims to show the extension and generalization abilities of RA-UNet and the segmentation on the whole tumor from the brain modality data. Thus, we merged NCR, NET, ED, and ET together to be the total tumor region. After that, the same strategy on patch extraction, which was used on liver tumor extraction, was applied on the BraTS2018 dataset. We extracted 400 tumor patches for each patient at a 64$\times$64$\times$64 resolution, and the whole training and validation datasets contain 114,000 patches. 

Compared to RA-UNet-\Rmnum{2}, we added more convolution filters for brain tumor segmentation in order to learn more tumor information. Detailed network setting is summarized in Table~\ref{table:RA-UNet brain params}. The other hyper parameter settings are the same with those in liver tumor segmentation. This version of RA-UNet has 12M parameters.

\begin{table}[]
\caption{Architecture of the proposed RA-UNET for brain tumor segmentation. Here [ ] denotes the long range connection; [ , ] denotes the concatenate operation; Conv means the convolution; Up stands for the up-sampling; Res denotes the residual block; and Att denotes the attention block}
\renewcommand\tabcolsep{6.8pt}      
\begin{tabular}{|c|c|c|c|c|}
\hline
Encoder & Output size                  & Decoder &                     & Output size                  \\ \hline
Input   & 64\textasciicircum{}3$\times$4  & Att1    & {[}Res4{]}, depth=0 & 4\textasciicircum{}3$\times$512  \\ \hline
Conv1   & 64\textasciicircum{}3$\times$32 & Res7    & {[}Up1, Att1{]}     & 4\textasciicircum{}3$\times$512  \\ \hline
Pooling & 32\textasciicircum{}3$\times$32 & Up2     &                     & 8\textasciicircum{}3$\times$512  \\ \hline
Res1    & 32\textasciicircum{}3$\times$64 & Att2    & {[}Res3{]}, depth=1 & 8\textasciicircum{}3$\times$256 \\ \hline
Pooling & 16\textasciicircum{}3$\times$64   & Res8    & {[}Up2, Att2{]}     & 8\textasciicircum{}3$\times$256 \\ \hline
Res2    & 16\textasciicircum{}3$\times$128   & Up3     &                     & 16\textasciicircum{}3$\times$256  \\ \hline
Pooling & 8\textasciicircum{}3$\times$128   & Att3    & {[}Res2{]}, depth=2 & 16\textasciicircum{}3$\times$128  \\ \hline
Res3    & 8\textasciicircum{}3$\times$256  & Res9    & {[}Up3, Att3{]}     & 16\textasciicircum{}3$\times$128  \\ \hline
Pooling & 4\textasciicircum{}3$\times$256  & Up4     &                     & 32\textasciicircum{}3$\times$128 \\ \hline
Res4    & 4\textasciicircum{}3$\times$512  & Att4    & {[}Res1{]}, depth=3 & 32\textasciicircum{}3$\times$64\\ \hline
Pooling & 2\textasciicircum{}3$\times$512   & Res10   & {[}Up4, Att4{]}     & 32\textasciicircum{}3$\times$64 \\ \hline
Res5    & 2\textasciicircum{}3$\times$512   & Up5     &                     & 64\textasciicircum{}3$\times$64 \\ \hline
Res6    & 2\textasciicircum{}3$\times$512   & Conv2   & {[}Up5, Conv1{]}    & 64\textasciicircum{}3$\times$32 \\ \hline
Up1     & 4\textasciicircum{}3$\times$512  & Conv3   &                     & 64\textasciicircum{}3$\times$1  \\ \hline
\end{tabular}
\label{table:RA-UNet brain params}
\end{table}

\begin{figure}
\centering
\includegraphics[scale=0.33]{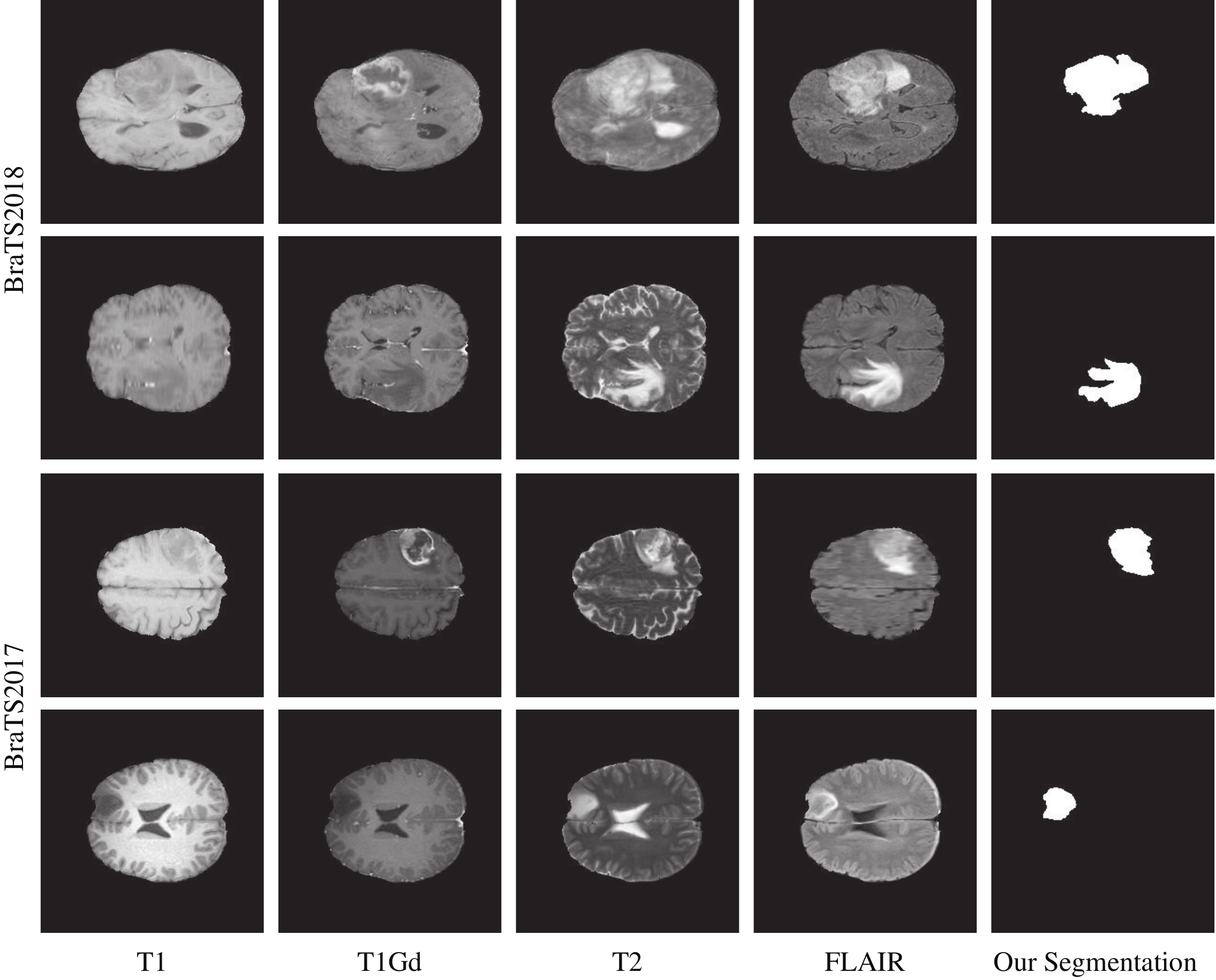}
\caption{Qualitative brain tumor segmentation results. RA-UNet has the ability of segmenting the whole tumor region. The first two rows come from BraTS2018 and the last two rows show the generalization of our model from BraTS2017.}
\label{fig:braTS results vis}
\end{figure}

The BraTS2017 and BraTS2018 leader board listed some state-of-the-art methods, and the Dice score of whole tumors reached 0.86$\sim$0.91. Table~\ref{table:braTS results comp} summarized several typical methods which perform well on the leader board. In Table~\ref{table:braTS results comp}, we can see that RA-UNet reaches the state-of-the-art performance and outperforms some other methods. The most important factor is that our model is a full 3D patch-based strategy, and it exhibits a high generalization ability on the BraTS2017 dataset without training prior. Typical segmentation slices of brain tumor are depicted in Fig.~\ref{fig:braTS results vis}, which indicates that RA-UNet is capable of segmenting brain tumor, and has a high extension ability.

\begin{table}[]
\caption{Performance of brain tumor segmentation on the BraTS dataset}
\renewcommand\tabcolsep{4pt} 
\renewcommand\arraystretch{1.3} 
\begin{tabular}{|c|c|c|c|c|c|}
\hline
Dataset       & Method           & DC              & Sensitivity     & Specificity     & Hausdorff95     \\ \hline
2018          & radiomics-miu    & 0.8764          & 0.8628          & 0.9950          & 4.9014          \\ \hline
2018          & GBMNet           & 0.8833          & 0.9340          & 0.9898          & 5.4614          \\ \hline
2018          & mmonteiro2       & 0.8709          & 0.8745          & 0.9932          & 5.7859          \\ \hline
2018          & UNeImage         & 0.8991          & 0.9101          & 0.9941          & 5.1043          \\ \hline
2018          & MIC-DKFZ         & 0.9125          & 0.9187          & 0.9954          & 4.2679          \\ \hline
\textbf{2018} & \textbf{RA-UNet} & \textbf{0.8912} & \textbf{0.8942} & \textbf{0.9938} & \textbf{5.8718} \\ \hline
2017          & BCVUniandes      & 0.8688          & 0.8420          & 0.9959          & 18.4569         \\ \hline
2017          & BRATZZ27         & 0.8800          & 0.8566          & 0.9960          & 5.7178          \\ \hline
2017          & CISA             & 0.8733          & 0.8548          & 0.9946          & 5.1805          \\ \hline
2017          & CMR              & 0.8569          & 0.8111          & 0.9968          & 5.8720          \\ \hline
2017          & MIC\_DKFZ        & 0.9026          & 0.9018          & 0.9957          & 6.7673          \\ \hline
2017          & Zhouch           & 0.9038          & 0.9032          & 0.9953          & 4.7447          \\ \hline
\textbf{2017} & \textbf{RA-UNet} & \textbf{0.8863} & \textbf{0.8697} & \textbf{0.9951} & \textbf{5.1112} \\ \hline
\end{tabular}
\label{table:braTS results comp}
\end{table}

\section{Conclusion}
\label{sec:conclusion}
To summarize our work, we have proposed an effective and efficient hybrid architecture for automatic extraction of liver and tumor from CT volumes. We introduce a new 3D residual attention-aware liver and tumor segmentation neural network named RA-UNet, which allows the extraction of 3D structures in a pixel-to-pixel fashion. The proposed network takes advantage of the strengths from the U-Net, the residual learning, and the attention residual mechanism. Firstly, attention-aware features change adaptively with the use of attention modules. Secondly, the residual blocks are stacked into our architecture which allows the architecture to go deeply and solve the gradient vanishing problem. Finally, the U-Net is used to capture multi-scale attention information and integrate low-level features with high-level features. To the best of our knowledge, this is the full 3D model and the first time that attention residual mechanism is implemented in the medical imaging tasks. Less parameters are trained by the attention residual mechanism. The effective system includes three stages: liver localization by a 2D RA-UNet, precise segmentation of liver, and tumor lesion by a 3D RA-UNet. More importantly, the trained network is a general segmentation model working on both the LiTS and the 3DIRCADb datasets.

Finally, we compared our approach with other methods including those from the LiTS challenge and those used on the 3DIRCADb dataset. In order to show the possibilities of extension for our model, we carried out brain tumor segmentation tasks on both BraTS2018 and BraTS2017 datasets. It indicates that our method achieved competitive results in liver tumor challenge, and exhibits high extension and generalization ability in brain tumor segmentation. In future work, we aim to further improve the architecture, making the architecture much more general to other tumor segmentation datasets and more flexible to common medical imaging tasks.

\section*{Acknowledgment}

This work is supported by the National Natural Science Foundation of China (Grant No. 61702361), the Science and Technology Program of Tianjin, China (Grant No. 16ZXHLGX00170), and the National Key Technology R\&D Program of China (Grant  No. 2015BAH52F00).

\ifCLASSOPTIONcaptionsoff
  \newpage
\fi

\bibliographystyle{IEEEtran}
\bibliography{bare_jrnl}

\end{document}